\documentclass[conference]{IEEEtran}
\IEEEoverridecommandlockouts

\usepackage{cite}
\usepackage{amsmath,amssymb,amsfonts}
\usepackage{algorithmic}
\usepackage{graphicx}
\usepackage{textcomp}
\usepackage{float}
\usepackage{multirow}
\usepackage[dvipsnames]{xcolor}
\usepackage{color,soul}
\usepackage{svg}

\renewcommand{\baselinestretch}{1}

\newcommand{\best}[1]{{\color{ForestGreen}{#1}}}

\def\BibTeX{{\rm B\kern-.05em{\sc i\kern-.025em b}\kern-.08em
    T\kern-.1667em\lower.7ex\hbox{E}\kern-.125emX}}
\begin{document}

%
%
\title{Hybrid ML for Lightweight Pre-Route Delay Estimation in Open-Source IC Design}


\author{\IEEEauthorblockN{Marvin Castro Castro}
\IEEEauthorblockA{
\textit{Universidad de Costa Rica}\\
marvin.castrocastro@ucr.ac.cr}
\and
\IEEEauthorblockN{Erick Carvajal Barboza}
\IEEEauthorblockA{
\textit{Universidad de Costa Rica}\\
erick.carvajalbarboza@ucr.ac.cr}
}

\maketitle

\begin{abstract}
Static Timing Analysis (STA) is a critical step in the design flow of digital integrated circuits, however,  obtaining accurate delay estimations can represent a challenge when limited information regarding physical design is available. In response, this work presents a hybrid and light-weight machine learning (ML) based approach that combines a decision tree with linear regression to improve pre-routing delay estimations generated by the open-source RTL-to-GDSII tool OpenLane. The proposed model achieves an 80\% reduction in error compared to OpenLane's estimates, demonstrates a 71\% improvement even without utilizing OpenLane-specific parameters. Overall, this method offers an alternative to traditional delay propagation techniques and more complex machine learning models that is not only accurate, but is also over 300 times smaller, 2 times faster and offers a higher explainability. 
\end{abstract}

\begin{IEEEkeywords}
Static Timing Analysis, Delay Prediction, Machine Learning, Decision Tree, Linear Regression, Ridge
\end{IEEEkeywords}

\section{Introduction}
One of the main reasons the VLSI design flow is often described as a feedback loop is the need for multiple iterations caused by timing violations identified during Static Timing Analysis (STA) in later stages of the design process. Even when designers carefully follow best practices during synthesis and placement, discrepancies between early estimations and post-layout results frequently require several re-optimization cycles. These iterative refinements make the process inefficient and time-consuming, as each iteration may involve logical and physical modifications such as gate resizing, logic restructuring, and buffer insertion to resolve timing issues~\cite{buffers}. As modern integrated circuits continue to scale in complexity and operate at increasingly higher frequencies, the number of these feedback loops tends to grow, significantly extending design turnaround time and impacting overall productivity.

To guarantee functionality under all conditions, STA tools often adopt a conservative approach known as pessimism. Pessimism consists of assuming that delays will be higher than the actual results, thereby ensuring that the circuit will still meet timing even under worst-case variations. While this approach can safeguard correctness and avoid potential field failures, it can also introduce new problems, such as suboptimal resource utilization, increased power consumption, and reduced space for additional logic~\cite{pessimism}. In other words, although pessimism helps prevent timing violations, it may penalize performance and area efficiency, leading to over-design. Delay estimation becomes particularly challenging during pre-routing stages, where limited physical information about the layout makes it difficult to accurately compute net delays. Without reliable interconnect parasitics, early estimates often diverge significantly from post-route values, making it difficult to perform accurate optimization before routing.

Most STA algorithms rely on a timing graph in which delays are calculated and propagated along all paths between startpoints and endpoints. These algorithms are generally scalable and fast, and they form the backbone of timing analysis in both academic and industrial tools. However, when applied in early design stages, such as before routing or detailed placement, they often lack sufficient physical information about the circuit, impairing the accuracy of their predictions. This limitation has motivated extensive research on alternative approaches to improve early timing estimation. Prior work~\cite{erick,lima1,lima2,lima3} has explored the use of machine learning (ML) techniques to address this challenge. By leveraging data from other completed designs or previous iterations, ML-based models can learn correlations between logical, structural, and partial physical features, aiding delay calculations or even generating independent predictions that complement traditional STA.

In this context, this work presents a hybrid model designed to predict path segment delays (comprising both gate and wire components) with improved accuracy during early design stages. The model is described as hybrid because it integrates two well-known, simple, and explainable learning algorithms: the decision tree and linear regression. Decision trees are particularly effective due to their ability to adapt to heterogeneous data distributions and their robustness to non-linear behavior. They are also resilient to outliers~\cite{dectree}, which are common in datasets derived from timing paths where unusual parasitic or fanout configurations may occur. However, decision trees have well-known limitations: they tend to overfit small datasets, can be sensitive to minor changes in the training data, and often become biased toward dominant features, reducing generalization capability. Moreover, their predictions are typically piecewise constant, which limits their interpolation ability between training samples.

On the other hand, linear regression remains one of the most widely used algorithms in learning-based prediction tasks due to its simplicity, interpretability, and computational efficiency. A well-regularized regression model can effectively generalize unseen data, providing smooth interpolation and capturing proportional relationships between timing features. Its primary weakness, however, lies in its assumption of global linearity, which makes it less suited for complex, highly non-linear datasets such as those found in pre-route timing prediction.

By combining both models, we aim to build a predictive framework that leverages the strengths of each while mitigating their individual weaknesses. The decision tree’s ability to partition the data space into distinct, homogeneous regions allows a linear regression model to be trained within each region of similar characteristics, ensuring that the regression operates on a subset of data where linearity is a reasonable assumption. This structure effectively enhances the model’s interpolation capabilities while preserving non-linear expressiveness. Additionally, the piecewise formulation provided by the decision tree makes it possible to handle non-linear dependencies between logical and physical features more flexibly, while maintaining computational efficiency. Both models are inherently explainable, which makes this hybrid approach transparent and easy to debug, an essential quality for adoption in design environments. The resulting model not only improves delay estimation accuracy but also provides designers with meaningful insights into the underlying factors influencing timing, thereby facilitating better-informed design decisions.

\section{Related Previous Work}

The prediction of Static Timing Analysis (STA) outcomes using learning-based models has been the subject of extensive research over the past decade. As the complexity of modern integrated circuits continues to increase, traditional analytical delay models often struggle to capture intricate interactions between logical, electrical, and physical features. Consequently, the use of data-driven techniques has become an attractive alternative to improve accuracy while maintaining computational efficiency. Early efforts in this direction focused on leveraging ensemble models such as Random Forests, which provide both robustness and interpretability.

In the work by Carvajal \textit{et al.}~\cite{erick}, a machine learning-based pre-routing model was implemented using a Random Forest structure, capable of reducing pessimism and yielding accurate timing predictions when compared against a commercial STA tool. Their approach demonstrated that even with limited physical information, it is possible to obtain reliable estimations of post-route timing behavior by learning statistical correlations from previous designs. Similarly, Kahng \textit{et al.}~\cite{bib2} introduced a methodology that combines machine learning with offset-based timing correlation to estimate wire delay and slew at the level of individual timing arcs. By integrating learned offsets into the traditional STA flow, their work significantly reduced endpoint slack estimation errors, illustrating how learning techniques can complement classical timing models without disrupting existing EDA toolchains.

More sophisticated models based on Neural Networks have also been explored to capture complex non-linear dependencies that simpler models might overlook. Studies such as~\cite{10627497,XTPRAGMA} employed Multi-Layer Perceptrons (MLPs) to predict path delays and analyze crosstalk effects, respectively. These works demonstrated the potential of deep architectures to automatically learn intricate timing relationships, achieving substantial improvements in accuracy over analytical baselines. However, the increased expressiveness of neural networks often comes at the cost of explainability and higher training data requirements, which may limit their adoption in practical design flows where transparency and runtime efficiency are critical. In addition, most of these studies have been tailored to commercial design environments, where proprietary tool data and closed-source timing engines provide the necessary detailed feature sets.

In contrast, recent research efforts have begun to focus on open-source Electronic Design Automation (EDA) tools, motivated by the need for reproducibility, accessibility, and community-driven innovation. The work by Sánchez \textit{et al.}~\cite{lima1} and Varela \textit{et al.}~\cite{lima2} are notable examples, as they apply Machine Learning techniques to improve the delay estimations produced by OpenLane, one of the most widely adopted open-source physical design frameworks. Using Random Forest and Gradient Boosted Decision Tree (GBDT) models, these studies achieved reductions of 67\% and 46\% in prediction error, respectively, compared to OpenLane’s default delay estimates. Their results confirmed that even with limited parasitic extraction and coarse-grained physical data, learning-based models can substantially improve pre-route delay prediction accuracy. Building on this line of work, another study by Varela \textit{et al.}~\cite{lima3} focused on predicting the delta between OpenLane’s pre-route and signoff delays, achieving up to 67\% less error relative to the open-source baseline. This delta-based formulation proved particularly effective, as it directly targeted the discrepancy between early and final timing stages, a persistent challenge in open-source STA.

Despite these advances, to the best of our knowledge, no prior work has employed a hybrid model similar to the one presented in this paper within the context of electronic design automation. While hybrid methods that combine a decision tree with multiple linear regressions applied at its leaf nodes have been successfully explored in other domains, their use in timing prediction or circuit design remains largely unexplored. For instance, Chen \textit{et al.}~\cite{bib4} utilized such a structure to forecast temperature variations up to seven days in advance, achieving improved interpretability and generalization compared to standalone models. Similarly, Azeez \textit{et al.}~\cite{bib5} applied comparable hybrid frameworks in diverse applications, including advertising analytics, sales forecasting, and the modeling of biological characteristics in animal species. These examples demonstrate the versatility and robustness of hybrid learning techniques when applied to heterogeneous datasets, suggesting that similar benefits could be achieved in EDA contexts.

Overall, the prior body of work highlights both the progress and limitations of existing approaches. While ensemble and neural models have demonstrated promising improvements in delay estimation accuracy, they often sacrifice explainability or require closed-source environments for deployment. This motivates the exploration of lightweight, transparent, and easily interpretable alternatives, such as the hybrid model proposed in this work, that can operate effectively within open-source frameworks like OpenLane while maintaining the level of insight and control expected by circuit designers.



\section{Methodology}

\subsection{Machine Learning Model Structure}

\begin{figure}[!htb]
\centering
\includegraphics[width=0.45\textwidth]{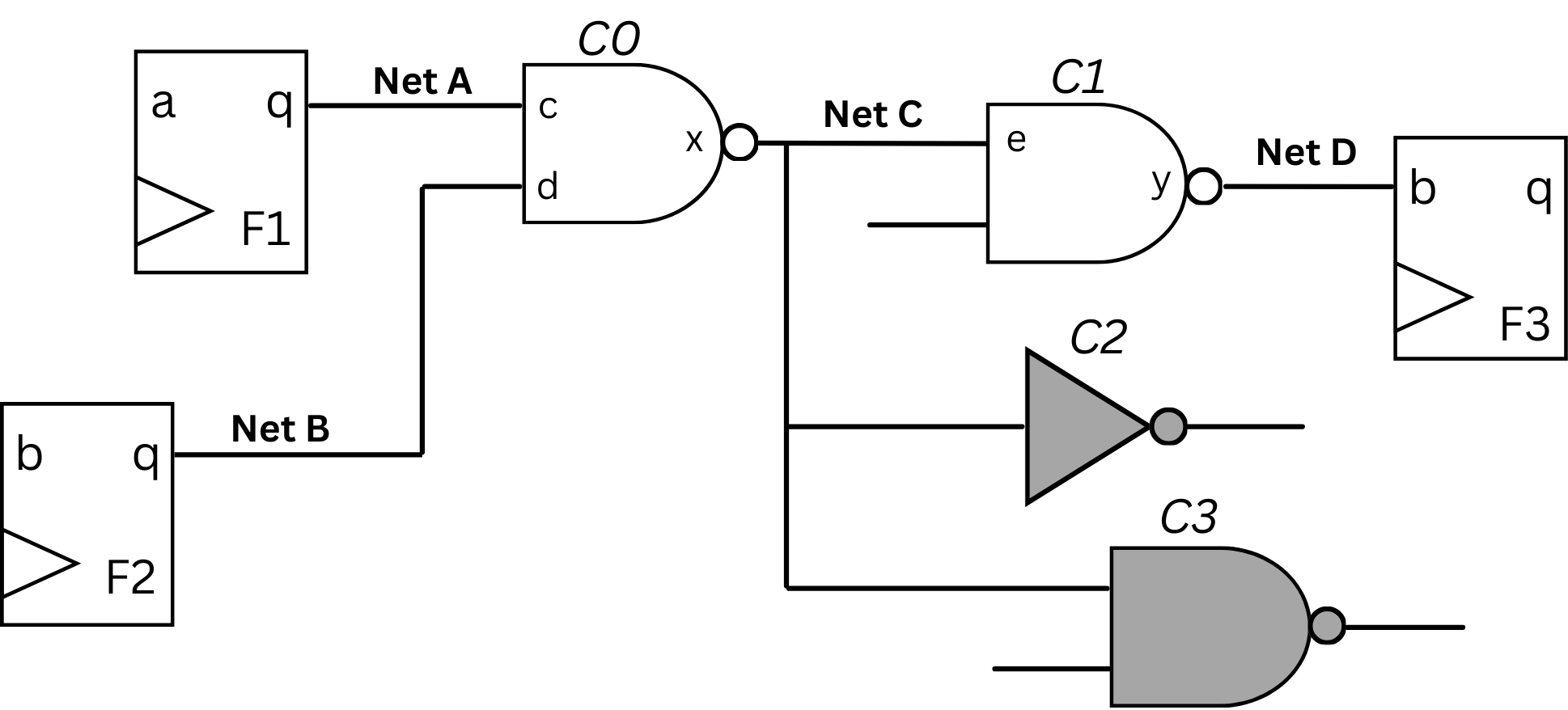} 
\caption{An example of a circuit}
\label{circuit_example}
\end{figure}

Our model predicts delay for individual nets using pre-routing information by analyzing each net’s driver and sink gates individually. For instance, in Figure \ref{circuit_example}, if the delay to sink e on Net C is to be estimated, capacitance, slew, and location information are obtained from gate C0, which serves as the driver, and gate C1, the sink for this path. The capacitance and location of both gates are considered. When estimating the delay to sink e, additional sinks, f and g, are incorporated as context sinks. These additional paths contribute to the load capacitance, while the locations of gates C2 and C3 also influence routing. 

\begin{figure}[!htb]
\centering
\includegraphics[width=0.45\textwidth]{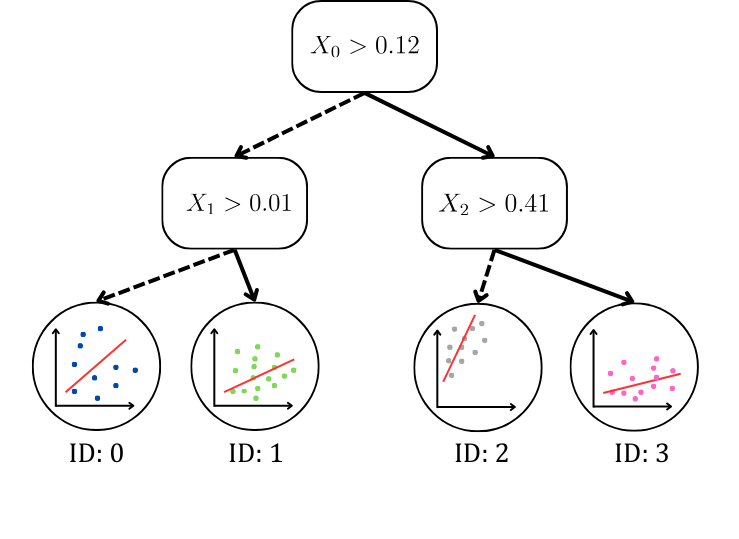} 
\caption{Hybrid model architecture }
\label{model_arch}
\end{figure}

Figure \ref{model_arch} describes the model architecture, where the top resembles a traditional decision tree, however each leaf node at the bottom is  linked to its own linear regression model, instead of a single constant value.

After the tree is constructed, for new data prediction, 
our model traverses the tree to determine the leaf node corresponding to each data sample and then uses the assigned linear regression model to estimate the delay.

\subsection{Model Features}
The model uses the logical and physical characteristics of each path segment as features, enabling it to make delay predictions based on those properties. These features include capacitance, slew, and distance for both driver and sink gates. Context-related parameters are incorporated to account for gates outside the path that may still influence the delay. 

\begin{itemize}
    \item \textbf{Fanout:} The number of gate inputs driven by a single gate's output. Higher fanout typically leads to increased load capacitance, which directly affects signal propagation delay.
    \item \textbf{Slew:} The signal transition time is also directly influenced by load capacitance; a higher slew rate correlates with an increase in propagation delay
    \item \textbf{Delay:} The delay parameter represents an approximation made during the pre-routing phase by the EDA tool, in our case, the open-source RTL-to-GDSII tool OpenLane.  
    \item \textbf{Driver and Sink Distance:} The distance between driver and sink reflects wire length which is proportional to our target variable. 
    \item \textbf{Driver and Sink Capacitance:} The driving strength of a gate correlates with its output capacitance, while sink capacitance represents the total load applied to the driver. The difference between these capacitances influences the net delay estimation. 
    \item \textbf{Driver and sink cell size:} Similarly to the driver and sink capacitances, this feature captures the difference between driving strength and load capacitance.

    \item \textbf{Context sinks' location and standard deviation} As previously noted, the presence of additional paths increase routing utilization, load capacitance and slew, resulting in higher delay. To account for these effects, the median location of all context sinks along the path is used as an approximation of the average distance from the driver to the context sinks. Additionally, the standard deviation of the context sinks location serves as a measure of spread, indicating how widely the sinks are distributed.  
\end{itemize}

\subsection{Data generation}

The feature data is obtained from pre-routing stages (\textit{e.g.} placement), while the data labels are obtained from signoff timing reports provided by OpenLane. Additional information, such as driver and sink locations, is gathered from pre-routing reports. The design flow was completed in OpenLane using the Skywater 130 nm Technology PDK to generate the data.
Data collection is performed by running the RTL-to-GDSII flow of various designs from benchmark circuits, including ISCAS-89~\cite{iscas}, OpenCores and examples provided by OpenLane. The Skywater 130 nm Technology PDK is used as the process node. Multiple runs from the same design are included, with variations related to placement, routing, clock frequency, pin layout, etc. 

\begin{figure}[!htb]
\centering
\includegraphics[width=0.35\textwidth]{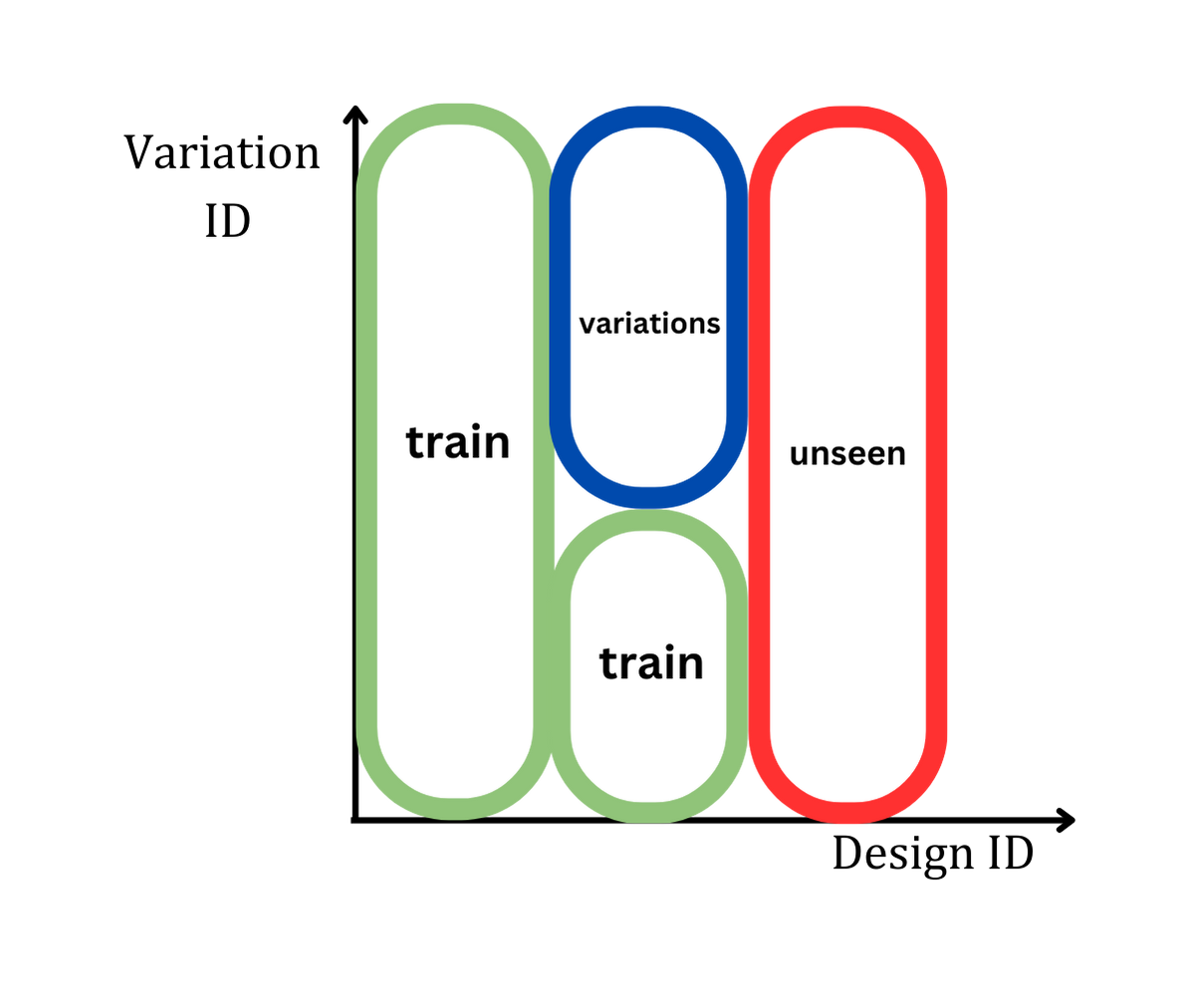} 
\caption{Datasets arrangement }
\label{data_cor}
\end{figure}

We divided the data into three datasets, with their relationships being illustrated in Figure \ref{data_cor}. The first, \textit{train}, is the largest of the three. It contains 8 different designs with up to 10 variations for each design. The \textit{variations} dataset contains a limited subset of some designs in \textit{train}, but with different variations. Specifically, circuits \verb|s38417| and \verb|s15850|. Finally, the \textit{unseen} dataset is independent of the other two, containing different designs along with their own variations. It includes data from circuits \verb|zipdiv| and \verb|picorv32|.


\subsection{Model Training and Selection}

For the selection of the model, numerous instances of the hybrid model were trained using various configurations of parameters on the training data. Different permutations of these parameter combinations were evaluated. The following variables were considered in 
the hybrid model:
\begin{itemize}
    \item \textbf{Data Scaling:} Including standardization, min-max feature scaling or no data scaling. Standardization places different variables on the same scale by making the values on each parameter have a mean equal to zero and unit-variance. The formula for standardization is given by
    \begin{equation*}
        x' = \frac{x-\bar{x}}{\sigma}
    \end{equation*}
    where $x$ is the original feature vector, $\bar{x}$ is the mean of the same feature vector, $\sigma$ is the standard deviation and $x'$ is the standardized value. 
    The min - max feature scaling brings all values of a parameter in the range [0, 1]. The formula for this transformation is given by
    \begin{equation*}
        x' = \frac{x- x_{min}}{x_{max} - x_{min}}
    \end{equation*}
    
    \item \textbf{Decision Tree Hyperparameters:} Maximum depth of the tree and maximum feature utilization both ranging from 5 to 15. 
    \item \textbf{Utilization of the median location of the context sinks}: Whether median location of context sink is used or not. 
    \item \textbf{Utilization of the standard deviation of the location of the sinks}: Whether standard deviation of the location of the sink is used or not. 
    \item \textbf{Utilization of the Distance parameter:} Two main scenarios were tested and compared. In the first, the model's features included both the $X$ and $Y$ coordinates of the driver and sink positions. In the second, these coordinates were replaced with a single feature representing the Euclidean distance between the driver and sink.
    \item \textbf{Type of linear regressor:} Ridge or Ordinary Least Squares (OLS) regression. 
\end{itemize}



\subsection{Machine Learning Algorithms}
This work utilizes three machine learning algorithms, each of which is described along with its corresponding cost function in this section. 

\begin{itemize}
    \item 
    \textbf{Decision Trees:} Decision trees predict outcomes by recursively partitioning the feature space~\cite{breiman2017classification}. Each node in the tree represents a split based on a feature, chosen to maximize the reduction in impurity, in our case, the Gini impurity. At each split, the tree chooses the feature and threshold that best separate the data to reduce variance in predictions. The cost function, often based on impurity (e.g., $\sum_{k=1}^K p_k(1 - p_k)$ for Gini), is minimized at each split to improve predictive accuracy. 
    
    \item \textbf{Linear Regression:} Corresponds to the ordinary least squares linear regression $y = \beta_0 x_0 + \beta_1 x_1 + ... + \beta_n x_n + \epsilon$. It minimizes the sum of the squared error between observed and predicted values with the cost function \ $\sum_{i=1}^n (y_i - \hat{y}_i)^2$~\cite{watson1967linear}. It offers no regularization, fitting all features equally. This makes it sensitive to outliers. 

    \item \textbf{Ridge Regression:} Utilizes the L2 regularization to provide an enhanced cost function that penalizes the size of weights $\beta_n$~\cite{hoerl1970ridge}. This helps mitigate possible overfitting or multicollinearity issues. The cost function is $ \sum_{i=1}^n (y_i - \hat{y}_i)^2 +  \lambda \sum_{j=1}^p \beta_j^2    $ 
    
\end{itemize}

\section{Results}
\subsection{Model Selection Results}
The best three results for the optimal model configurations are shown in Table \ref{tab:modelselec}. The first entry achieved the smallest RMSE on the \textit{unseen} dataset. The second entry displays the smallest RMSE for the \textit{variations} dataset, demonstrating the model's ability to perform well even with a reduced tree depth. The third row presents the configuration for the smallest combined error for both datasets, and finally the last row presents the estimations of OpenLane during pre-route stages. All configurations used Ridge as the linear regressor. Notably, the hybrid model achieved up to five times less error compared to OpenLane, and the correlation coefficient remained above 0.9 for all variations of the model. 

Additionally, Table~\ref{tab:modelselec} shows the size in MB of three different fitted models, with an average size of 10.86MB. In contrast, the models presented in \cite{lima1,lima2,lima3} range from 4GB to 6GB, making the hybrid model several times lighter than its Random Forest and Gradient Boosted Tree counterparts.

\begin{table*}[!htb]
\centering
\caption{Model Selection Results}
\label{tab:modelselec}
\resizebox{0.99\textwidth}{!}{%
\begin{tabular}{ccccc|cc|cc}
\hline
\multirow{2}{*}{\textbf{Data Scaling}} & \multirow{2}{*}{\textbf{\begin{tabular}[c]{@{}c@{}}Context\\ Features\end{tabular}}} & \multirow{2}{*}{\textbf{\begin{tabular}[c]{@{}c@{}}Distance\\ Parameter\end{tabular}}} & \multirow{2}{*}{\textbf{Tree Depth}} & \multirow{2}{*}{\textbf{\begin{tabular}[c]{@{}c@{}}.pkl Size\\ (MB)\end{tabular}}} & \multicolumn{2}{c|}{\textbf{Variations}} & \multicolumn{2}{c}{\textbf{Unseen}} \\ \cline{6-9} 
 &  &  &  &  & \textbf{RMSE} & \textbf{Correlation} & \textbf{RMSE} & \textbf{Correlation} \\ \hline
Normalized & No & Yes & 10 & 9.83 & 0.5977 & 0.93 & 0.8193 & 0.91 \\
Normalized & Yes & Yes & 5 & 12.94 & 0.5465 & 0.94 & 0.8489 & 0.90 \\
No & No & Yes & 13 & 9.83 & 0.5942 & 0.93 & 0.8212 & 0.91 \\ \hline
\multicolumn{5}{c|}{\textbf{OpenLane Pre-Route Estimation}} & 2.503 & 0.428 & 1.199 & 0.704 \\ \hline
\end{tabular}%
}
\end{table*}

\begin{table*}[!htb]
\centering
\caption{Detailed Delay Prediction Results for the Hybrid Model and the Open Source Tool}
\label{tab:detaileddelay}
\resizebox{0.9\textwidth}{!}{%
\begin{tabular}{c|cc|cc|cc}
\hline
 & \multicolumn{2}{c|}{\textbf{Hybrid Model w/ Delay}} & \multicolumn{2}{c|}{\textbf{Hybrid Model wo/ Delay}} & \multicolumn{2}{c}{\textbf{OpenLane pre-route}} \\ \hline
\textbf{Circuit} & \textbf{RMSE} & \textbf{Correlation} & \textbf{RMSE} & \textbf{Correlation} & \textbf{RMSE} & \textbf{Correlation} \\ \hline
picorv32 & 0.879 & 0.901 & 1.276 & 0.775 & 1.216 & 0.696 \\
zipdiv & 0.859 & 0.899 & 1.238 & 0.777 & 1.057 & 0.807 \\
s38417 & 0.663 & 0.924 & 0.973 & 0.824 & 2.730 & 0.413 \\
s15850 & 0.656 & 0.922 & 0.947 & 0.826 & 0.629 & 0.333 \\ \hline
\end{tabular}%
}
\end{table*}

\begin{table*}[!htb]
\centering
\caption{Detailed Delay Comparison Between the Hybrid Model and Other Models}
\label{tab:delaycomparison}
\resizebox{\textwidth}{!}{%
\begin{tabular}{c|c|ccc|ccc}
\hline
\multicolumn{1}{l|}{} & \multicolumn{1}{l|}{\multirow{2}{*}{\textbf{\begin{tabular}[c]{@{}l@{}}Pre-Route\\ OpenLane\end{tabular}}}} & \multicolumn{3}{c|}{\textbf{Estimations w/ Delay}} & \multicolumn{3}{c}{\textbf{Estimations wo/ Delay}} \\
\textbf{} & \multicolumn{1}{l|}{} & \multicolumn{1}{c|}{\textbf{Full Estimation \cite{lima1}}} & \multicolumn{1}{c|}{\textbf{Delta Estimation \cite{lima3}}} & \textbf{Ours} & \multicolumn{1}{c|}{\textbf{Full Estimation \cite{lima2}}} & \multicolumn{1}{c|}{\textbf{Delta Estimation \cite{lima3}}} & \textbf{Ours} \\ \hline
s15850 & 0.629 & \multicolumn{1}{c|}{0.563} & \multicolumn{1}{c|}{\best{0.522}} & 0.656 & \multicolumn{1}{c|}{\best{0.679}} & \multicolumn{1}{c|}{0.848} & 0.947 \\
s38417 & 2.73 & \multicolumn{1}{c|}{0.849} & \multicolumn{1}{c|}{0.835} & \best{0.663} & \multicolumn{1}{c|}{1.2176} & \multicolumn{1}{c|}{\best{0.550}} & 0.973 \\
picorv32 & 1.216 & \multicolumn{1}{c|}{1.008} & \multicolumn{1}{c|}{1.024} & \best{0.879} & \multicolumn{1}{c|}{1.3001} & \multicolumn{1}{c|}{\best{1.076}} & 1.276 \\
zipdiv & 1.057 & \multicolumn{1}{c|}{1.052} & \multicolumn{1}{c|}{1.031} & \best{0.859} & \multicolumn{1}{c|}{1.3975} & \multicolumn{1}{c|}{1.040} & \best{1.238} \\ \hline
\end{tabular}%
}
\end{table*}

\subsection{Delay prediction results}

Results in this section use the model in the third row of Table~\ref{tab:modelselec}, which uses no data scaling and no context features, as this configuration provided the best outcomes for both test sets. Table \ref{tab:detaileddelay} offers detailed results of error for each circuit in the testing pool. On average, the hybrid model reduces error by up to 65\% compared to OpenLane across all circuits. Notably, for the third and fourth circuits, the model achieves significantly lower error rates, with reductions of 77\% for \verb|s38417| and 80\% for \verb|s15850|. These superior results may be attributed to the increased similarity between these circuits and the training data. Additionally, Table \ref{tab:detaileddelay} includes results for the hybrid model when the pre-route delay estimation by OpenLane is used as input feature for the model, as well as when it is not used, this is done in order to compare the results with the one obtained in prior work~\cite{lima1,lima2}. In this configuration, the model demonstrates a substantial error reduction of 69\% exclusively for the \textit{variations} dataset.

\subsection{Delay prediction and model comparison}

Table \ref{tab:delaycomparison} compares the results of the work by Sanchez \textit{et al.}~\cite{lima1}, as well as Varela \textit{et al.}~\cite{lima2,lima3}, against the ones obtained by this work. While some of these~\cite{lima1,lima2} employ Random Forest models for delay prediction, others~\cite{lima3} predict the difference between the pre-route delay estimated by OpenLane and the signoff delay, and is therefore categorized as a delta prediction. Since some of these works use the pre-route delay estimation as part of the input features to the model, while some do not, we categorize the results according to the usage of the value.

When the pre-route delay estimation of OpenLane is used as an input feature to the ML models, the hybrid model has the highest accuracy in the \textit{unseen} dataset (that is \verb|picorv32| and \verb|zipdiv|), as well as the lowest for \verb|s38417|, while staying close to the error of the best performing for \verb|s15850|.

When pre-route estimation is not an input feature, the delta-based model~\cite{lima3} continues to be the best performing overall, although this work outperforms it in the case of \verb|zipdiv|. 

Overall, the hybrid model yields a lower error in 3 out of the 4 evaluated circuits, achieving up to a 28\% reduction in RMSE for \verb|s38417| and an average reduction of 21.76\% compared to the delay estimations from \cite{lima1}. The only circuit where the hybrid model is outperformed is \verb|s15850|, where it shows a 25.7\% higher error than the best-performing model.

In terms of the inference time, the reported inference time for Gradient Boosted Decision Trees is about 366ms, while for the Hybrid Model is about 170ms. Is important to consider that the code that we used to implement the hybrid model has opportunities to be optimized further, while the code for the Gradient Boosted Trees are already highly optimized since it is implemented using Scikit-Learn library. 







\section{Conclusion}
To the best of our knowledge, this work represents the first integration of a hybrid model for delay prediction in integrated circuits, combining the strengths of decision trees and linear regression within a unified framework. The resulting model is lightweight, computationally efficient, and demonstrates measurable improvements in several delay prediction scenarios when compared to both the open-source tool OpenLane and other standalone machine learning approaches with similar objectives. By effectively partitioning the feature space and applying localized regression models, the hybrid structure enhances interpolation capability while preserving explainability—an essential characteristic for design engineers seeking to understand and trust automated predictions.

A distinctive advantage of the proposed approach lies in its interpretability. Each prediction can be decomposed into two intuitive components: the sequence of decisions that guided the sample through the tree, and the explicit linear formula applied at the corresponding leaf node. This transparency not only facilitates debugging and model validation but also provides valuable insights into which design features contribute most significantly to delay variations. In this sense, the model does not act as a “black box” but rather as an assistive analytical tool that can complement traditional STA and help designers make informed optimization choices earlier in the design flow.

Despite the encouraging results obtained in this study, the authors acknowledge several avenues for further improvement. The current dataset, while sufficient to demonstrate feasibility, could be expanded to encompass a broader variety of circuits, process nodes, and design conditions. A larger and more diverse dataset would likely enhance the model’s generalization capability and enable a more comprehensive evaluation of its performance limits. Moreover, future research could investigate the incorporation of additional features, such as routing congestion metrics, switching activity, or coupling capacitance estimates, to further refine the delay prediction accuracy.

\bibliographystyle{IEEEtran}
\bibliography{refs}

\vspace{12pt}

\end{document}